\documentclass{article} 
\usepackage[preprint]{colm2026_conference}

\usepackage{microtype}
\usepackage{hyperref}
\usepackage{url}
\usepackage{booktabs}
\usepackage{graphicx}   
\usepackage{tcolorbox}
\usepackage{verbatim}
\tcbuselibrary{listings, breakable}
\usepackage{listings}
\usepackage{multirow}

\usepackage{lineno}

\definecolor{darkblue}{rgb}{0, 0, 0.5}
\hypersetup{colorlinks=true, citecolor=darkblue, linkcolor=darkblue, urlcolor=darkblue}

\title{Representation Alignment as a Bottleneck in LLM-Based Retrosynthesis Planning}

\author{
Hyunwoo Yoo$^{1}$ \quad
Cassie Huang$^{1}$ \quad
Haebin Shin$^{2}$ \quad
Li Zhang$^{1}$ \quad
Gail L. Rosen$^{1}$ \\
$^{1}$Drexel University \\
$^{2}$University of Michigan \\
\texttt{\{hty23,ch3535,hz466,glr26\}@drexel.edu}
\quad
\texttt{haebin@umich.edu}
}

\begin{document}

\ifcolmsubmission
\linenumbers
\fi

\maketitle

\begin{abstract}

While LLMs show promise in general reasoning, symbolic planning in chemistry remains a bottleneck. Direct ''SMILES-to-PDDL'' attempts fail because they force models to juggle chemical analysis and planning-language structuring simultaneously.
We hypothesize that this failure stems from a lack of intermediate abstractions rather than insufficient model capacity. By decomposing retrosynthesis into molecule mapping, reaction mapping, and PDDL generation, we achieve high success rates where end-to-end approaches fail. This provides evidence that a primary bottleneck lies in representation alignment rather than raw model capacity.
Our structural analysis demonstrates that intermediate representations are essential in retrosynthesis planning, highlighting the importance of representation-centric design in future systems.
\end{abstract}

\section{Introduction}


Large Language Models (LLMs) have emerged as powerful general-purpose reasoners~\citep{chen2021codex, wei2022chain}, yet they continue to struggle with structured symbolic planning tasks such as chemical retrosynthesis. Formulating retrosynthesis as a direct translation from molecular representations such as the Simplified Molecular Input Line Entry System (SMILES) to symbolic planning languages like the Planning Domain Definition Language (PDDL) ~\citep{mcdermott1998pddl, fox2003pddl21} introduces significant challenges.
This task requires the model to simultaneously perform chemical structure interpretation, reaction-level structuring, and formalization into a symbolic planning language~\citep{garrett2020pddlstream, silver2021symbolic}—a complex reasoning process that is difficult to resolve in a single step.
Importantly, representing retrosynthesis in PDDL enables the generation of executable and interpretable plans, where each action explicitly encodes reaction steps and dependencies, allowing verification by external planners.

In this work, rather than attributing these failures to insufficient model capacity, we ask a more fundamental question: \textit{\textbf{Do LLMs truly lack planning ability, or do they fail during the transformation between heterogeneous representations?}}
Our analysis reveals an intriguing paradox. While LLMs achieve strong performance on individual subtasks—such as molecule mapping, reaction mapping, and PDDL generation—they fail completely when these processes are integrated into a single end-to-end transformation~\citep{bubeck2023sparks,zhou2023least}. In particular, under the direct SMILES-to-PDDL setting, all evaluated models exhibit a 0\% planning success rate.
These findings suggest that the failure of LLMs does not stem from a lack of reasoning capability, but rather from representation misalignment between unstructured chemical representations and symbolic planning formalisms~\citep{yao2022react,chen2023program}. In other words, while models are capable of performing individual transformations and structuring operations, they struggle to consistently bridge heterogeneous representation spaces.

To systematically validate this hypothesis, we propose an analytical framework that decomposes retrosynthesis planning into staged subtasks: molecule mapping, reaction mapping, and PDDL generation. Each stage incrementally transforms unstructured inputs into structured symbolic representations, and we isolate and evaluate performance at each stage to precisely identify where failures occur.

Experimental results show that, when structured intermediate representations are provided, most strong models achieve high planning success rates, indicating that downstream symbolic planning itself is already a largely solvable problem. In contrast, when attempting to solve the task end-to-end without such intermediate representations, all models fail, supporting the claim that the primary bottleneck lies not in model scale or architecture, but in representation alignment.

The contributions of this work are threefold. 
\begin{itemize}
    \item We redefine the failure of LLM-based retrosynthesis planning as a problem of representation misalignment rather than insufficient model capability, explaining the discrepancy between strong subtask performance and complete end-to-end failure.
    \item We propose a staged analytical framework and corresponding benchmarks that decompose retrosynthesis planning into interpretable intermediate transformations, enabling systematic analysis of failure across representation levels.
    \item We demonstrate that planning performance is governed more by representation alignment than by the underlying model itself.
\end{itemize}

Our findings highlight the limitations of end-to-end approaches and underscore the importance of intermediate symbolic representations in the design of future LLM-based planning systems.

\begin{figure}[t!]
    \centering
    \includegraphics[width=\linewidth]{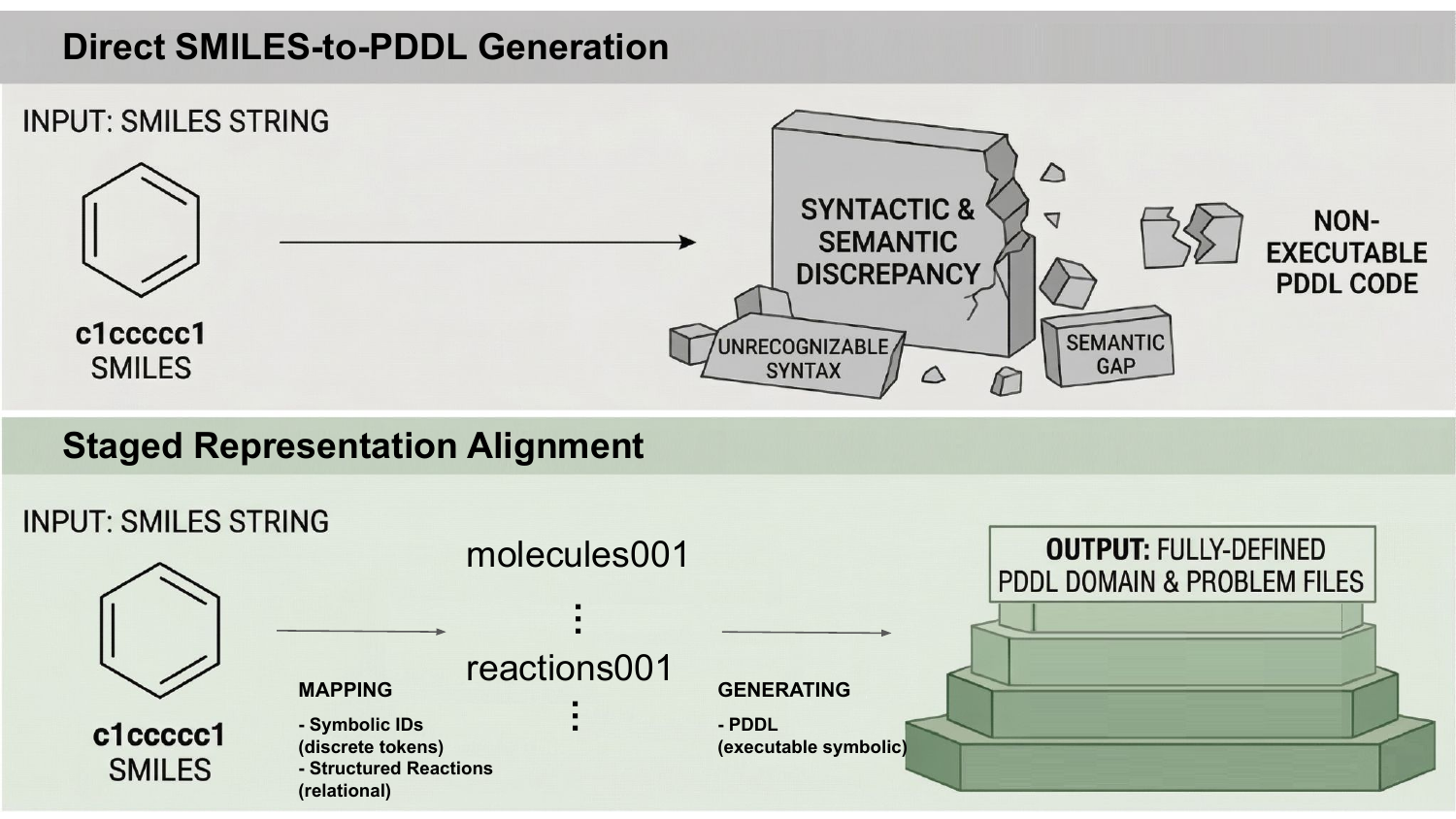}
    \caption{
        \textbf{Representation alignment as the bottleneck in retrosynthesis planning.}
        \textbf{Top:} Direct SMILES-to-PDDL generation requires transforming unstructured molecular inputs into executable symbolic plans in a single step.
        \textbf{Bottom:} Our staged approach introduces intermediate representations (symbolic molecule identifiers and structured reaction mappings) before generating PDDL.
    }    
    \label{fig:overview}
\end{figure}

\section{Related Work}

\subsection{LLMs for Planning}

Recent research has actively explored the use of large language models for planning problems~\citep{valmeekam2022planbench,valmeekam2023planning,ahn2022saycan,huang2022innermonologue,lin2023gplanet}. These works typically evaluate LLMs’ planning capabilities by generating action sequences from natural language instructions or by leveraging intermediate outputs in the form of code or programs to facilitate the planning process~\citep{liang2022codeaspolicies,yao2023react,gao2022pal,chen2022pot}. Some studies further propose solving classical planning problems directly with language models or improving performance by integrating them with external planners~\citep{liu2023llmp,valmeekam2023planning, helmert2006fastdownward}. However, the majority of prior work focuses on natural language-based planning or standardized benchmark environments, and does not systematically analyze failures arising in the transformation from unstructured inputs to executable symbolic representations such as PDDL~\citep{mcdermott1998pddl, fox2003pddl21}. In this work, we take a different perspective by centering our analysis on representation transitions in the context of retrosynthesis planning.

\subsection{LLMs in Chemistry and Retrosynthesis}

In the domain of chemistry, the use of LLMs and foundation models has rapidly expanded, covering a wide range of tasks such as molecular representation understanding, reaction prediction, synthesis pathway recommendation~\citep{shen2021automation, striethkalthoff2024retrosynthesis}, and chemical question answering~\citep{dealmeida2019synthetic,struble2020ai,bran2024chemcrow,zhang2024chemllmchemicallargelanguage}. Retrosynthesis, in particular, is a fundamental problem that involves inferring possible precursors from a target molecule, and recent efforts have explored using language models to generate reaction rules or propose synthesis pathways~\citep{schwaller2019moleculartransformer,ucak2022retrotrae,zhong2023graph2edits,han2024editretro}. However, these studies primarily focus on reaction prediction or pathway generation itself, and relatively little attention has been given to formalizing the outputs into symbolic planning representations~\citep{tamari2021process, odonoghue2023bioplanner, anhel2023lap} that can be directly executed by planners~\citep{liu2023llmp}. In contrast, our work treats retrosynthesis not merely as a generation problem, but as an executable symbolic planning problem, and distinguishes itself by analyzing at which stage LLMs fail in this process.

\subsection{Intermediate Representations and Modular Reasoning}

Decomposing complex reasoning problems into multiple stages with intermediate representations has long been proposed as an important design principle~\citep{wei2022chain,gao2022pal,chen2022pot,yao2023react,schick2023toolformer}. In recent LLM research, various forms of intermediate representations—such as chain-of-thought, program-of-thought, tool use, and structured decoding—have been shown to contribute to performance improvements~\citep{wei2022chain,chen2022pot,gao2022pal,yao2023react,schick2023toolformer}. Some works further demonstrate that modularizing problems to isolate errors at each substage can lead to more robust reasoning~\citep{gao2022pal,yao2023react,liu2023llmp}. However, these ideas have primarily been studied in the context of natural language reasoning or code generation, and there has been limited investigation into the role of representation alignment in settings that require bridging unstructured chemical inputs with symbolic planning formalisms, such as retrosynthesis planning. In this work, we introduce a staged analytical framework consisting of molecule mapping, reaction mapping, and PDDL generation, and quantitatively demonstrate the role of intermediate symbolic abstraction in enabling executable planning.

\section{From Molecules to Plans: A Structured Transformation Framework}

\subsection{Overview}

In this work, to analyze where large language models (LLMs) fail in retrosynthesis planning, we decompose the transformation process from unstructured chemical representations to executable symbolic plans into a sequence of stages. Each stage operates at a different level of representation, where the input is progressively structured and ultimately translated into a planning language.
Specifically, we organize the overall process into four stages. First, molecule mapping converts symbolic molecular identifiers into molecular structure representations. Second, reaction mapping structures reaction-level information into organized reactant–product representations. Third, PDDL generation formalizes these structured reactions into an executable planning format. Finally, planning execution verifies whether the generated representation can produce a valid synthesis pathway.
This decomposition is not merely intended to simplify the problem, but to precisely identify failures that arise during transitions between different representations. Through this, we aim to uncover bottlenecks that are not visible in end-to-end settings and to develop a structural understanding of failure modes in LLM-based planning.

\subsection{Molecule Mapping}

The first stage maps symbolic molecule identifiers to their corresponding molecular structure representations. In our setting, the input consists of symbolic identifiers referring to molecules, and the model is required to generate the corresponding SMILES strings for each identifier.
Formally, given a set of molecules {mi}, the model generates a set of SMILES strings {si} corresponding to each molecule. Evaluation is conducted based on exact-match accuracy between identifiers and SMILES, as well as the rate of missing or invalid outputs.
This stage measures how reliably symbolic references are grounded into actual molecular structures, and serves as the foundation for subsequent reaction-level structuring and planning representation generation.

\subsection{Reaction Mapping}

The second stage transforms reaction information into structured reactant–product representations. In this setting, the model takes as input the molecular identifiers involved in each reaction and constructs the corresponding sets of reactants and products.
Formally, each reaction is represented as a tuple (R, P), where R and P denote the sets of reactants and products, respectively. Evaluation is based on reaction-level exact match, measuring structural correctness including missing, duplicated, or mismatched reactants or products.
This stage aligns molecule-level information into reaction-level structures and forms the intermediate representation that enables subsequent conversion into PDDL.

\subsection{PDDL Generation}

In the third stage, structured reaction information is converted into the PDDL. Given a set of reactions, the model generates both the planning domain and problem definitions.
The domain file represents each reaction as an action, where preconditions and effects symbolically define the structure of the reaction. The problem file specifies the initial and goal states, defining the planning instance to be solved.
This stage is evaluated along three dimensions. First, syntactic validity measures whether the generated PDDL conforms to correct syntax. Second, structural completeness evaluates whether required elements such as actions, predicates, domain, and problem components are properly included. Third, semantic consistency measures whether each action accurately reflects the underlying reactant–product relationships.
This stage constitutes the core step that transforms structured chemical information into executable symbolic planning representations.

\subsection{Planning Execution}

In the final stage, we verify whether the generated PDDL is actually executable. To this end, we employ an external classical planner, the Fast Downward system~\citep{helmert2006fastdownward}, to search for synthesis pathways based on the domain and problem definitions.
Evaluation is conducted using two metrics. The solve rate measures the proportion of instances for which the planner successfully generates a valid plan, while path accuracy measures whether the generated reaction pathway matches the ground-truth solution.
This stage is crucial in that it ensures the outputs of the LLM are not merely syntactically plausible, but lead to truly executable plans.

\begin{figure}[t]
    \centering
    \includegraphics[width=\linewidth]{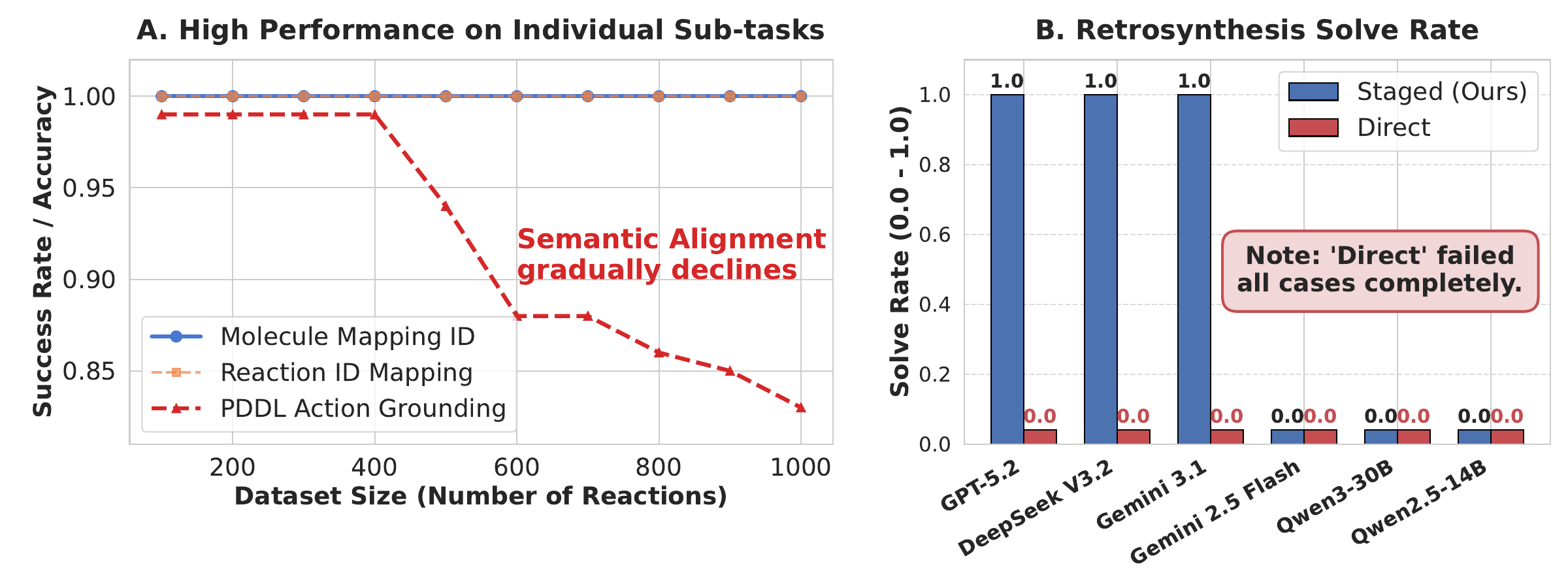}
    \caption{
    Comparison between decomposed and end-to-end planning regimes.
    (A) Sub-task performance across dataset sizes, including molecule mapping, reaction mapping, and PDDL grounding.
    (B) Planning success (solve rate) comparing direct end-to-end generation and staged decomposition.
    }
    \label{fig:comparison_decomposed_end2end}
\end{figure}


\subsection{Evaluation Protocol}

We evaluate LLM-based retrosynthesis planning under different settings to analyze performance variations across representation levels.
First, in the end-to-end setting, the model directly generates PDDL from SMILES inputs, treating the entire process as a single-step generation problem. This represents the most direct formulation, where the model must perform the transformation from unstructured chemical inputs to executable planning representations in one step.
In contrast, in the decomposed setting, we evaluate each stage independently with oracle intermediate representations to measure conditional performance of specific transformation steps. This setup enables disentangling upstream errors from downstream generation capabilities.
We additionally evaluate a non-oracle chained setting in which model-generated outputs are passed sequentially between stages.
Importantly, the goal of this study is not to propose a fully automated pipeline, but to identify at which representation stage failures occur in the process of transforming unstructured chemical inputs into executable symbolic plans. Through this comparative analysis, we precisely locate the primary bottlenecks in LLM-based planning.

\section{RetroPlan-Bench: Symbolic Retrosynthesis Planning Benchmark}

\subsection{Construction Overview}
In this work, to systematically evaluate the symbolic planning capabilities of LLMs, we propose RetroPlan-Bench, which reconstructs existing retrosynthesis datasets into an executable symbolic planning benchmark.
This benchmark is built upon 368 multi-step synthesis pathways from the READRetro~\citep{kim2024readretro}, and transforms each pathway into a multi-level representation aligned across molecule-level, reaction-level, and planning-level abstractions.
Through this, we establish an evaluation framework that enables quantitative decomposition and analysis of the entire process from unstructured chemical information to executable symbolic plans.

\subsection{Symbolic Transformation Pipeline}

RetroPlan-Bench does not simply utilize raw chemical data;
instead, it applies a three-stage transformation pipeline designed to analyze the representation alignment capabilities of LLMs.
\paragraph{Symbolic Grounding.}
All molecules (SMILES) are replaced with unique symbolic identifiers.
This encourages the model to focus on relational and compositional structures between symbols rather than the intrinsic complexity of chemical structures.
\paragraph{Reaction Structuring.}
Each synthesis pathway is decomposed into (reactant, product) pairs,
which are then aligned into structured reaction representations with unique reaction identifiers.
\paragraph{Coverage Filtering.}
Only pathways that can be fully represented within a predefined reaction set are retained,
ensuring that failures in planning arise from the model’s representation transformation capabilities rather than missing data.
This pipeline goes beyond simple preprocessing,
providing a controlled symbolic abstraction that explicitly exposes transformations across heterogeneous representation spaces.

\subsection{Multi-scale Planning Regimes}

To analyze problem complexity and scaling behavior of models,
we construct planning environments of varying sizes based on the number of reactions.
To evaluate actual planning performance, we select pathways such that the number of unique reactions is 100, 200, 300, and 400 as problem sets.
Pathways are chosen to ensure a balanced distribution of difficulty, considering both path length and branching structure.
To analyze limitations in the PDDL generation stage,
we augment the base reaction sets with additional reactions to construct large-scale planning domains containing up to 1000 actions.
This setting is designed to evaluate the model’s ability to handle long contexts and maintain structural consistency.

\subsection{Evaluation Protocol}

For each dataset configuration, we evaluate performance along four distinct dimensions: 1) Molecule Mapping Accuracy, which measures the alignment between symbolic identifiers and molecular structures; 2) Reaction Consistency, which assesses the correctness of reactant–product structures; 3) PDDL Validity and Grounding, which evaluates both syntactic validity and the preservation of reaction semantics; and 4) Planning Success, which measures executability via an external planner as well as the accuracy of generated pathways. This decomposed evaluation enables precise identification of representation alignment failures that cannot be observed from end-to-end performance alone.

\begin{figure}[t]
    \centering
    \includegraphics[width=\linewidth]{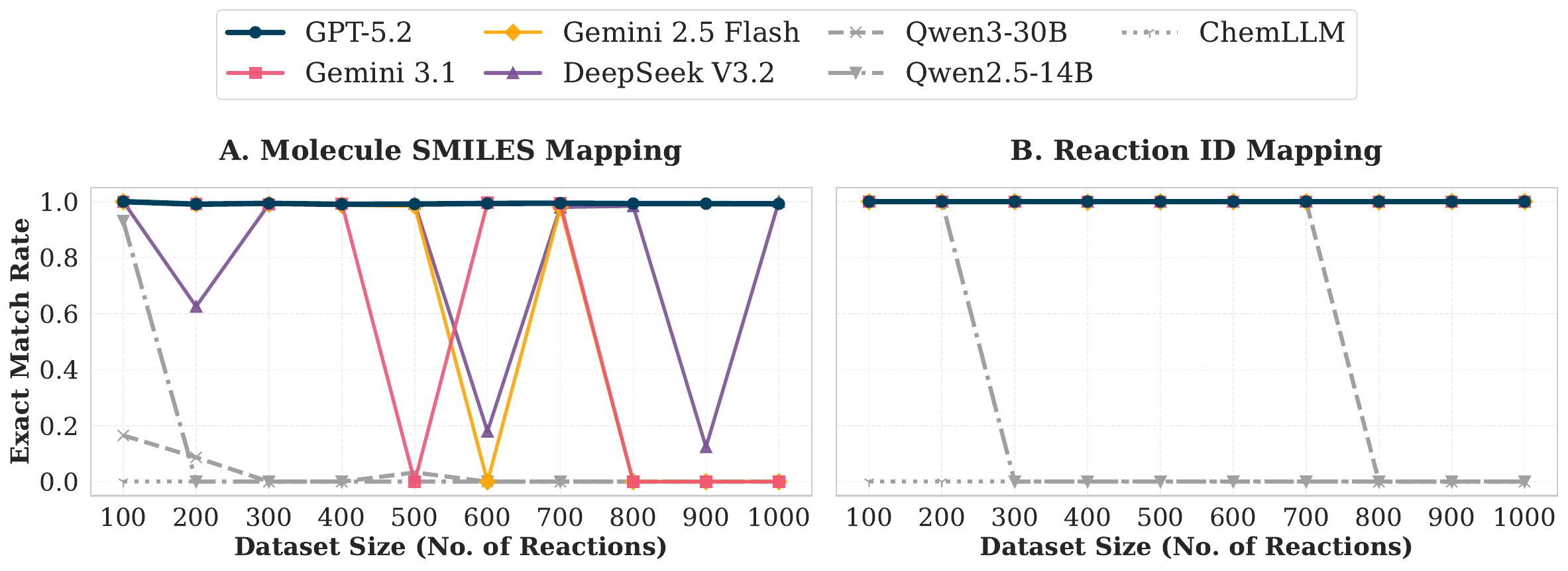}
    \caption{
    Analysis of representation alignment across abstraction levels.
    (A) Molecule SMILES mapping accuracy, measuring exact-match grounding of symbolic identifiers.
    (B) Reaction ID mapping accuracy, evaluating structural consistency of reactant–product relationships.
    }    
    \label{fig:representation_alignment_analysis}
\end{figure}

\begin{table*}[t]
\centering
\small
\begin{tabular}{lccccc}
\toprule
\multirow{2}{*}{\textbf{Model}} & \multicolumn{2}{c}{\textbf{Molecule Mapping}} & \textbf{Reaction Mapping} & \multicolumn{2}{c}{\textbf{PDDL Grounding}} \\
\cmidrule(r){2-3} \cmidrule(r){4-4} \cmidrule(l){5-6}
& \textbf{Molecule ID} & \textbf{SMILES} & \textbf{Coverage} & \textbf{Domain} & \textbf{Problem} \\
\midrule
GPT-5.2             & 1.0000 & 0.9932 & 1.0000 & 0.9303 & 1.0000\\
DeepSeek V3.2       & 0.9992 & 0.7870 & 1.0000 & 0.9300 & 1.0000\\
Gemini 3.1          & 0.6000 & 0.5964 & 1.0000 & 0.8402 & 1.0000\\
Gemini 2.5 Flash    & 0.6000 & 0.5933 & 0.9996 & 0.8351 & 1.0000\\
Qwen3-30B-Thinking  & 0.2976 & 0.0285 & 0.7000 & 0.4670 & 0.9128\\
Qwen2.5-14B-Instruct& 0.0991 & 0.0930 & 0.2000 & 0.4041 & 0.3991\\
ChemLLM             & 0.0000 & 0.0000 & 0.0000 & 0.0000 & 0.0000\\
\bottomrule
\end{tabular}
\caption{
Performance comparison across staged subtasks in retrosynthesis planning (averaged).
Molecule Mapping includes \textbf{Molecule ID} (identifier consistency) and \textbf{SMILES} (exact-match string accuracy).
\textbf{Reaction Mapping} (Coverage) measures correctness of reactant–product structure.
\textbf{PDDL Grounding} includes \textbf{Domain} (action generation) and \textbf{Problem} (initial and goal specification).
}
\label{tab:overall_summary}
\end{table*}


\section{Experiments and Results}
In this section, we analyze how LLMs perform in generating executable symbolic plans from unstructured chemical representations, and identify at which stages failures occur. In particular, by comparing performance between individual subtasks and the end-to-end setting, we aim to precisely diagnose the source of failure.

\subsection{Strong Performance on Individual Sub-tasks}
We first evaluate each subtask that composes retrosynthesis planning—molecule mapping, reaction mapping, and PDDL generation—independently.
Experimental results (Table~\ref{tab:overall_summary}) show that GPT-5.2~\citep{openai2023gpt4, openai2025gpt52}, DeepSeek V3.2~\citep{deepseekai2025deepseekv32pushingfrontieropen}, and Gemini-family models~\citep{geminiteam2023gemini} achieve near-perfect performance across most settings. In molecule mapping, identifier exact match converges to 1.0 across all regimes, and reaction mapping similarly maintains an accuracy of 1.0 in nearly all configurations.
A similar trend is observed in the PDDL generation stage. Strong models consistently maintain high syntactic validity and structural completeness for both domain and problem definitions, producing outputs that are reliably executable by planners.
In contrast, Qwen-family models (Qwen3-30B-Thinking~\citep{yang2025qwen3technicalreport}, Qwen2.5-14B-Instruct~\citep{qwen2025qwen25technicalreport}) and ChemLLM~\citep{zhang2024chemllmchemicallargelanguage} exhibit some level of performance in earlier stages, but show a rapid decline in coverage and validity as the task progresses toward PDDL generation. In particular, domain generation suffers from a sharp breakdown in structural completeness as the number of reactions increases.
These results indicate that, for sufficiently strong models, individual transformation steps are already largely solved.

\subsection{Failure of End-to-End Generation}
Next, we evaluate an end-to-end setting in which the same models including GPT-5.2, DeepSeek V3.2~\citep{deepseekai2025deepseekv32pushingfrontieropen}, Gemini 3.1, Gemini 2.5 Flash, 
Qwen3-30B-Thinking~\citep{yang2025qwen3technicalreport}, Qwen2.5-14B-Instruct~\citep{qwen2025qwen25technicalreport}, and ChemLLM~\citep{zhang2024chemllmchemicallargelanguage} are tasked with directly generating PDDL domain and problem definitions from SMILES inputs.
The results show that, across all models, both solve rate and path accuracy drop to 0, with not a single valid synthesis pathway generated (Figure~\ref{fig:comparison_decomposed_end2end}), despite their strong performance on individual subtasks (Table~\ref{tab:overall_summary}).
This indicates that models which perform well on individual subtasks completely fail when the entire transformation process is integrated into a single step.
In other words, despite possessing the capability to perform each operation independently, models struggle to compose these operations into a coherent transformation pipeline.

\subsection{Effect of Intermediate Representations}
To analyze the cause of this performance degradation, we introduce a setting in which intermediate representations are provided in a staged manner.
Under this setup, models with strong capacity such as GPT-5.2, DeepSeek V3.2, and Gemini 3.1 recover planning performance, achieving a solve rate of 1.0 and high path accuracy (Figure~\ref{fig:comparison_decomposed_end2end}).
In contrast, smaller models including Gemini 2.5 Flash, Qwen3-30B-Thinking, Qwen2.5-14B-Instruct, and ChemLLM fail to produce executable plans even with structured inputs.
This demonstrates that the same models that fail in the end-to-end setting can successfully generate executable plans when given structured intermediate representations.
Thus, these results suggest that a primary challenge lies in transforming input representations into appropriate symbolic forms.
To test whether this improvement is an artifact of oracle intermediate representations, we additionally evaluate a non-oracle chained setting in which each stage receives the model-generated output from the previous stage. GPT-5.2 and DeepSeek-V3.2 retain the same performance as in the oracle-staged setting (Solve Rate = 1.0, Path Accuracy = 0.986).

\subsection{Scaling Behavior and Failure Modes}
As we increase dataset scale and analyze performance at each stage, we observe not only differences across models but also distinct failure patterns (Figure~\ref{fig:representation_alignment_analysis}).
GPT-5.2 and DeepSeek V3.2 maintain stable domain coverage and syntactic validity across all regimes (Table~\ref{tab:overall_summary}); however, as the number of reactions increases, action grounding accuracy gradually declines. This suggests that while structural form is preserved, semantic consistency may degrade with scale.
Gemini-family models (Gemini 3.1 and Gemini 2.5 Flash) generally maintain high performance, but exhibit sharp drops in domain validity or grounding at specific scales, followed by recovery at larger sizes. This pattern indicates instability in long-horizon structural generation.
In contrast, Qwen-family models (Qwen3-30B-Thinking and Qwen2.5-14B-Instruct) show rapid degradation in coverage and validity even at relatively small scales, and in some cases fail to produce valid symbolic structures altogether. Semantic grounding also deteriorates significantly.
These results suggest that differences across models extend beyond raw accuracy, reflecting their ability to maintain stable structured representations.

\subsection{Planning Performance}
Finally, we evaluate actual planning performance by executing a planner on the generated PDDL.
In settings with structured intermediate representations, GPT-5.2, DeepSeek V3.2~\citep{deepseekai2025deepseekv32pushingfrontieropen}, and Gemini 3.1 all achieve a solve rate of 1.0 and high path accuracy, indicating their ability to reliably generate executable symbolic plans.
In contrast, Qwen-family models~\citep{qwen2025qwen25technicalreport, yang2025qwen3technicalreport}, Gemini 2.5 Flash, and ChemLLM~\citep{zhang2024chemllmchemicallargelanguage} achieve 0 in both solve rate and path accuracy. 
A closer inspection reveals that ChemLLM's failure is dominated by malformed structured outputs (e.g., invalid JSON and non-executable PDDL), rather than purely chemistry-specific reasoning errors (see Appendix~\ref{app:chemllm_failures}).
Notably, Gemini 2.5 Flash demonstrates strong performance on some subtasks, yet fails at the final planning stage, highlighting that accuracy at individual stages does not necessarily translate to executability.
Meanwhile, in the end-to-end setting where PDDL is directly generated from SMILES, all models fail to produce valid plans. This highlights a clear gap between individual capabilities and the ability to generate executable plans in an integrated setting.
Additional detailed results and per-scale analyses are provided in Appendix~\ref{sec:appendix_detailed_results}. 
Representative failure cases of direct SMILES-to-PDDL generation are discussed in Appendix~\ref{sec:fail_case}.

\section{Discussion}

\subsection{Representation Alignment vs. Reasoning Capability}

The central question of this work is as follows:
\textit{Do LLM failures stem from a lack of planning capability itself, or do they arise from the transformation between heterogeneous representations?}
Our experimental results strongly support the latter.
Across all subtasks—molecule mapping, reaction mapping, and PDDL generation—strong models achieve near-perfect performance, indicating that individual structuring and transformation operations are well within their capabilities. In contrast, the fact that the same models completely fail under the end-to-end setting suggests that the core issue does not lie in reasoning capability itself.
Moreover, the recovery of planning performance when intermediate representations are provided indicates that, while models possess the necessary operations, they struggle to consistently compose them across different representation spaces.
From this perspective, failures in retrosynthesis planning are better characterized not as reasoning failures, but as cross-representation composition failures.

\subsection{Why End-to-End Transformation Fails}

First, the input (SMILES) and output (PDDL) exist at different levels of abstraction and structure, and the transformation between them requires not merely conversion but re-structuring across representation spaces. 
Second, such transformations require explicit structural anchoring at intermediate stages. However, in the end-to-end setting, these intermediate representations are only implicitly handled within the model, increasing the likelihood that information becomes progressively distorted or lost across stages.
Third, as observed in our experiments, some models maintain syntactic structure while gradually losing semantic grounding, or exhibit abrupt structural collapse at certain scales. This suggests that satisfying multiple constraints simultaneously over long generation processes is inherently challenging.

\subsection{The Role of Intermediate Symbolic Representations}

Intermediate symbolic representations play a crucial role in mitigating these issues.
Our results show that when intermediate stages such as molecule mapping and reaction mapping are explicitly separated, models can satisfy structural constraints at each stage independently, leading to substantial improvements in overall planning performance.
Furthermore, these findings suggest that although LLMs are capable of handling multiple levels of abstraction internally, explicit structural decomposition is necessary to consistently externalize these representations into coherent outputs.

\subsection{Implications for LLM-based Planning Systems}

The findings of this work provide several important implications for the design of LLM-based planning systems.
First, end-to-end approaches that treat the generation of executable plans from unstructured inputs as a single-step problem may face fundamental limitations.
Second, staged designs centered around intermediate symbolic representations are not merely an engineering choice, but can be a key determinant of performance.
Third, planning performance is influenced less by model scale or general reasoning ability, and more by how consistently alignment across heterogeneous representations can be maintained.

\subsection{Limitations and Future Directions}
This study focuses on the specific domain of retrosynthesis planning, and further investigation is needed to determine whether similar phenomena generalize to other planning problems.
While our primary stage-wise analysis uses oracle intermediate representations, the additional non-oracle chained evaluation remains limited to the same retrosynthesis benchmark and does not establish domain-general robustness.
Future work should aim to generalize the representation alignment problem and explore model architectures or training strategies that can effectively address it.

\section{Conclusion}

In this work, we analyze the failure of LLMs in retrosynthesis planning and reinterpret it not as a limitation of reasoning capability, but as a problem of representation alignment.
By decomposing the planning process into staged subtasks and analyzing performance at each stage, we empirically identify a gap between individual operational capabilities and overall problem-solving ability through comparison with the end-to-end setting. In particular, the finding that all models fail under the end-to-end setting, while achieving strong planning performance when provided with structured intermediate representations, clearly demonstrates the importance of intermediate representations in generating executable symbolic plans.
These results highlight the limitations of end-to-end approaches in LLM-based planning systems and emphasize that representation-centric structural design can play a crucial role in future research.


\section*{Acknowledgments}
This work is supported in part by funds from the National Science Foundation (NSF: \# 2107108).


\bibliography{colm2026_conference}
\bibliographystyle{colm2026_conference}

\appendix

\section{Failure Cases in Direct SMILES-to-PDDL Generation}
\label{sec:fail_case}

To further analyze why direct SMILES-to-PDDL prompting fails, we inspect representative outputs from strong language models under the end-to-end setting. Rather than exhaustively listing all malformed generations, we highlight a small number of qualitatively distinct failure modes that consistently appear across models.

These examples demonstrate that failures are not limited to minor syntax errors. Instead, models struggle to preserve raw SMILES strings as stable, planner-compatible symbols, resulting in outputs that are syntactically malformed, semantically corrupted, or non-executable. This observation supports our main claim that the primary bottleneck lies in cross-representation alignment rather than downstream planning itself.

\begin{figure*}[t]
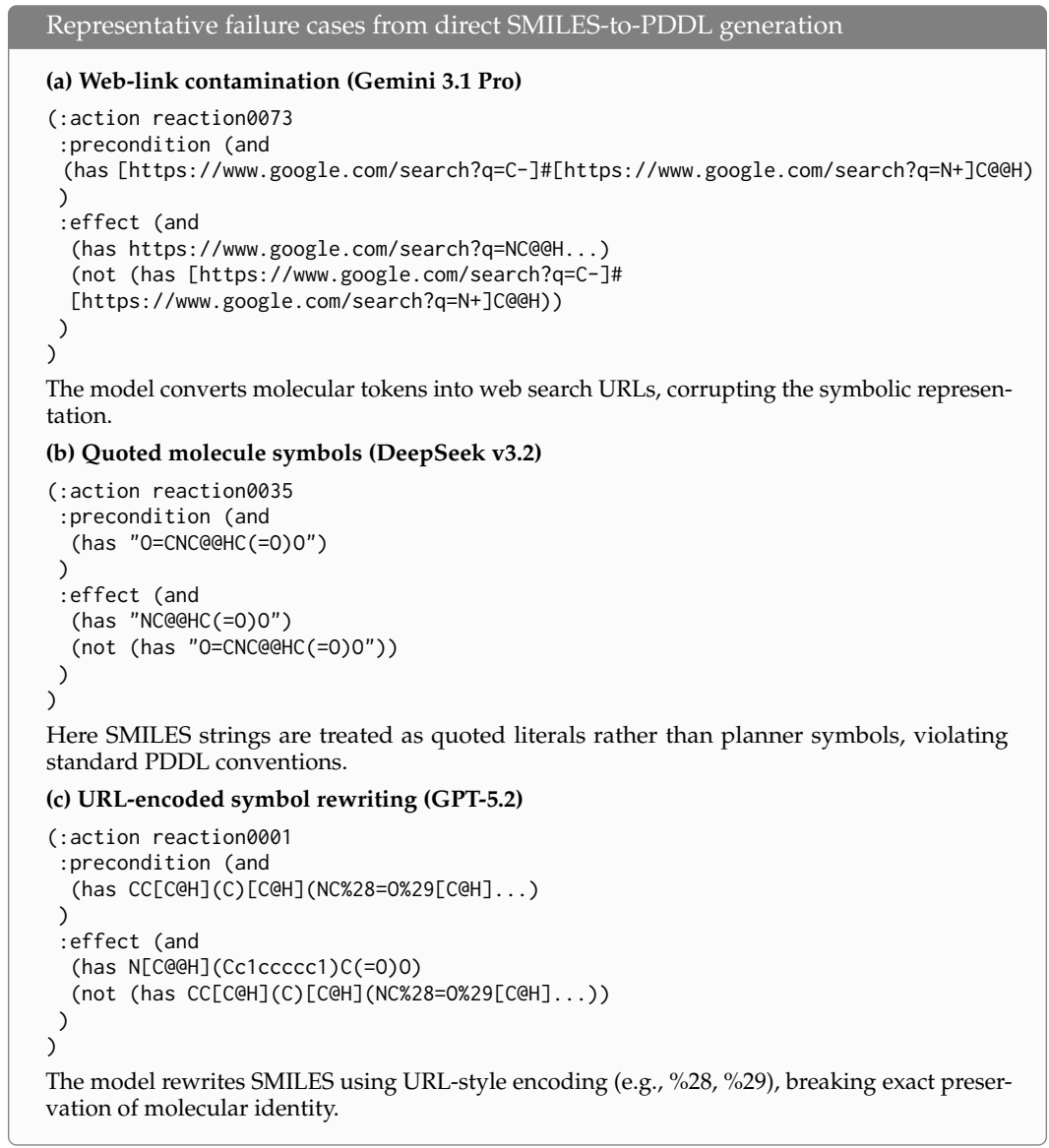

\centering
\begin{tcolorbox}[
    colback=gray!3,
    colframe=black!50,
    width=\textwidth,
    boxrule=0.5pt,
    title={Representative failure cases from direct SMILES-to-PDDL generation}
]
\small

\textbf{(a) Web-link contamination (Gemini 3.1 Pro)}
\begin{verbatim}
(:action reaction0073
 :precondition (and
  (has [https://www.google.com/search?q=C-]#[https://www.google.com/search?q=N+]C@@H)
 )
 :effect (and
  (has https://www.google.com/search?q=NC@@H...)
  (not (has [https://www.google.com/search?q=C-]#  
  [https://www.google.com/search?q=N+]C@@H))
 )
)
\end{verbatim}
The model converts molecular tokens into web search URLs, corrupting the symbolic representation.

\vspace{0.5em}
\textbf{(b) Quoted molecule symbols (DeepSeek v3.2)}
\begin{verbatim}
(:action reaction0035
 :precondition (and
  (has "O=CNC@@HC(=O)O")
 )
 :effect (and
  (has "NC@@HC(=O)O")
  (not (has "O=CNC@@HC(=O)O"))
 )
)
\end{verbatim}
Here SMILES strings are treated as quoted literals rather than planner symbols, violating standard PDDL conventions.

\vspace{0.5em}
\textbf{(c) URL-encoded symbol rewriting (GPT-5.2)}
\begin{verbatim}
(:action reaction0001
 :precondition (and
  (has CC[C@H](C)[C@H](NC%28=O%29[C@H]...)
 )
 :effect (and
  (has N[C@@H](Cc1ccccc1)C(=O)O)
  (not (has CC[C@H](C)[C@H](NC%28=O%29[C@H]...))
 )
)
\end{verbatim}
The model rewrites SMILES using URL-style encoding (e.g., \%28, \%29), breaking exact preservation of molecular identity.

\end{tcolorbox}
\caption{Representative failure modes in direct SMILES-to-PDDL generation. Although surface forms differ across models, all cases reflect a shared issue: raw SMILES strings are not reliably maintained as executable symbolic entities.}
\label{fig:direct_smiles_failure_cases}
\end{figure*}

Across models, these failures are superficially different but structurally related. Some models reinterpret SMILES tokens as web-searchable text, others convert them into quoted string literals, and others apply escaping or encoding schemes. In all cases, the model fails to preserve molecular representations as stable symbolic objects that can be consumed by a classical planner.

This suggests that direct SMILES-to-PDDL generation imposes a compounded burden: the model must simultaneously maintain chemical string identity, impose reaction-level structure, and satisfy symbolic planning syntax within a single generation process. Without explicit intermediate representations, these constraints are not consistently satisfied, leading to systematic breakdowns in executability.

\subsection{Failure Under Explicit Output Constraints}

To rule out the possibility that the observed failures are due to underspecified output formatting, we additionally impose explicit constraints on how molecular symbols should be represented. In particular, we instruct the model to either preserve SMILES strings exactly or convert them into deterministic, PDDL-safe symbols (e.g., via sanitization or mapping rules).

Despite these explicit constraints, the same failure patterns persist. Models continue to produce malformed outputs, including quoted string literals, web-link-like tokens, fragmented molecule symbols, and inconsistent rewritings of the same SMILES string. In some cases, models even generate degenerate actions where preconditions and effects collapse to identical or trivial forms, indicating a loss of semantic grounding.

These results suggest that the failure cannot be attributed solely to prompt ambiguity or insufficient specification of output format. Rather, even when the target representation is explicitly constrained, models fail to consistently maintain molecular identity and structural relationships across the generation process.

This further supports our central claim that the primary bottleneck lies in representation alignment. The model must simultaneously preserve chemical string identity, enforce reaction-level structure, and satisfy symbolic planning syntax. The inability to jointly satisfy these constraints—despite explicit instruction—indicates a fundamental limitation in composing transformations across heterogeneous representation spaces.

\subsection{Failure Persists Under Sanitized Symbol Mapping}

One may argue that the direct SMILES-to-PDDL setting is excessively brittle because raw SMILES strings contain characters that are not naturally suited for planner-facing symbolic outputs. To control for this possibility, we additionally provide an explicit deterministic mapping from each SMILES string to a PDDL-safe molecule symbol and instruct the model to use only these sanitized symbols.

However, failures persist even under this setting. Although the raw SMILES strings are no longer required in the final PDDL, models still exhibit several characteristic breakdowns. First, some models truncate or fragment long mapped symbols, producing incomplete identifiers that no longer match the provided mapping. Second, some outputs collapse distinct molecules into overly simplified or repeated symbols, causing semantic loss. Third, some generations remain partially malformed or are cut off before the full domain is completed. These errors indicate that the difficulty is not limited to handling special characters in SMILES, but extends to maintaining long-range symbolic consistency even when a planner-safe vocabulary is explicitly supplied.

\begin{figure*}[t]
\centering
\begin{tcolorbox}[
    colback=gray!3,
    colframe=black!50,
    width=\textwidth,
    boxrule=0.5pt,
    title={Representative failure cases under sanitized symbol mapping}
]
\small

\textbf{(a) Symbol truncation / fragmentation (GPT-5.2)}
\begin{verbatim}
(:action reaction0001
 :precondition (and
  (has mol_CC_C_H_C_C_H_NC_O_C_H_Cc1ccc_O_cc1_NC_O__7a59f0fb6a)
 )
 :effect (and
  (has mol_N_C_H_Cc1ccccc1_C_O_O_6b44e61ea4)
  (not (has mol_CC_C_H_C_C_H_NC_O_C_H_Cc1ccc_O_cc1_NC_O__7a59f0fb6a))
 )
)
\end{verbatim}
Although closer to the intended format, the generated symbol sequence is still vulnerable to truncation and loss of readability when very long mapped identifiers are used.

\vspace{0.5em}
\textbf{(b) Partial completion / malformed continuation (Gemini 3.1 Pro)}
\begin{verbatim}
mol_Cc1cc_O_cc2c1C_O_CC_O_Cc1cc_O_cc_O_o1_O2_60acf903d3)
(not (has mol_Cc1cc_O_c2c_c1_Cc1cccc_O_c1C2_O_987c495792)))
)

(:action reaction0003
 :precondition (and (has mol_NCCc1ccc_O_cc1_65bbbc584d))
 :effect (and (has mol_N_C_H_Cc1ccc_O_cc1_C_O_O_3c2bc25bc4)
              (not (has mol_NCCc1ccc_O_cc1_65bbbc584d)))
)
\end{verbatim}
The output resumes from an already malformed fragment, indicating instability in maintaining structured action boundaries even when valid symbolic mappings are supplied.

\vspace{0.5em}
\textbf{(c) Semantic collapse to repeated generic forms (DeepSeek v3.2)}
\begin{verbatim}
(:action reaction0055
 :precondition (and
  (has "NC@@HC(=O)O")
 )
 :effect (and
  (has "NC@@HC(=O)O")
  (not (has "NC@@HC(=O)O"))
 )
)

(:action reaction0056
 :precondition (and
  (has "NC@@HC(=O)O")
 )
 :effect (and
  (has "NC@@HC(=O)O")
\end{verbatim}
Even under constrained prompting, distinct reactions collapse into the same repeated molecular form, producing semantically degenerate actions.

\end{tcolorbox}
\caption{Failure modes under explicit sanitized-symbol mapping. Supplying planner-safe symbols reduces the burden of raw SMILES formatting, but does not resolve the deeper problem of maintaining stable symbolic grounding and structurally consistent long-form generation.}
\label{fig:sanitized_mapping_failures}
\end{figure*}

These examples strengthen the interpretation that end-to-end failure is not caused solely by the lexical surface form of SMILES. If the issue were merely the presence of special characters, then replacing molecules with deterministic planner-safe symbols should largely resolve the problem. Instead, the remaining failures show that models also struggle with consistent symbol reuse, action-level semantic grounding, and stable long-horizon structured generation.

In this sense, sanitized mapping improves the surface representation but does not eliminate the underlying cross-representation bottleneck. The model is still required to preserve molecule identity, align reactants and products correctly, and serialize them into executable symbolic plans without losing structural consistency. The persistence of failure under this control setting therefore provides additional evidence that the main limitation lies in representation alignment rather than output formatting alone.

\subsection{Failure Modes of ChemLLM in Structured Symbolic Generation}
\label{app:chemllm_failures}

Although ChemLLM is a chemistry-specialized language model, it failed across all evaluated stages in our benchmark. Inspection of raw generations suggests that its failure is not primarily due to subtle chemistry-specific reasoning errors, but rather due to instability in constrained symbolic serialization. Across molecule mapping, reaction mapping, and PDDL generation, ChemLLM frequently fails to preserve exact identifiers, satisfy rigid output schemas, or maintain the formal structure required for executable symbolic representations.

\begin{figure*}[t]
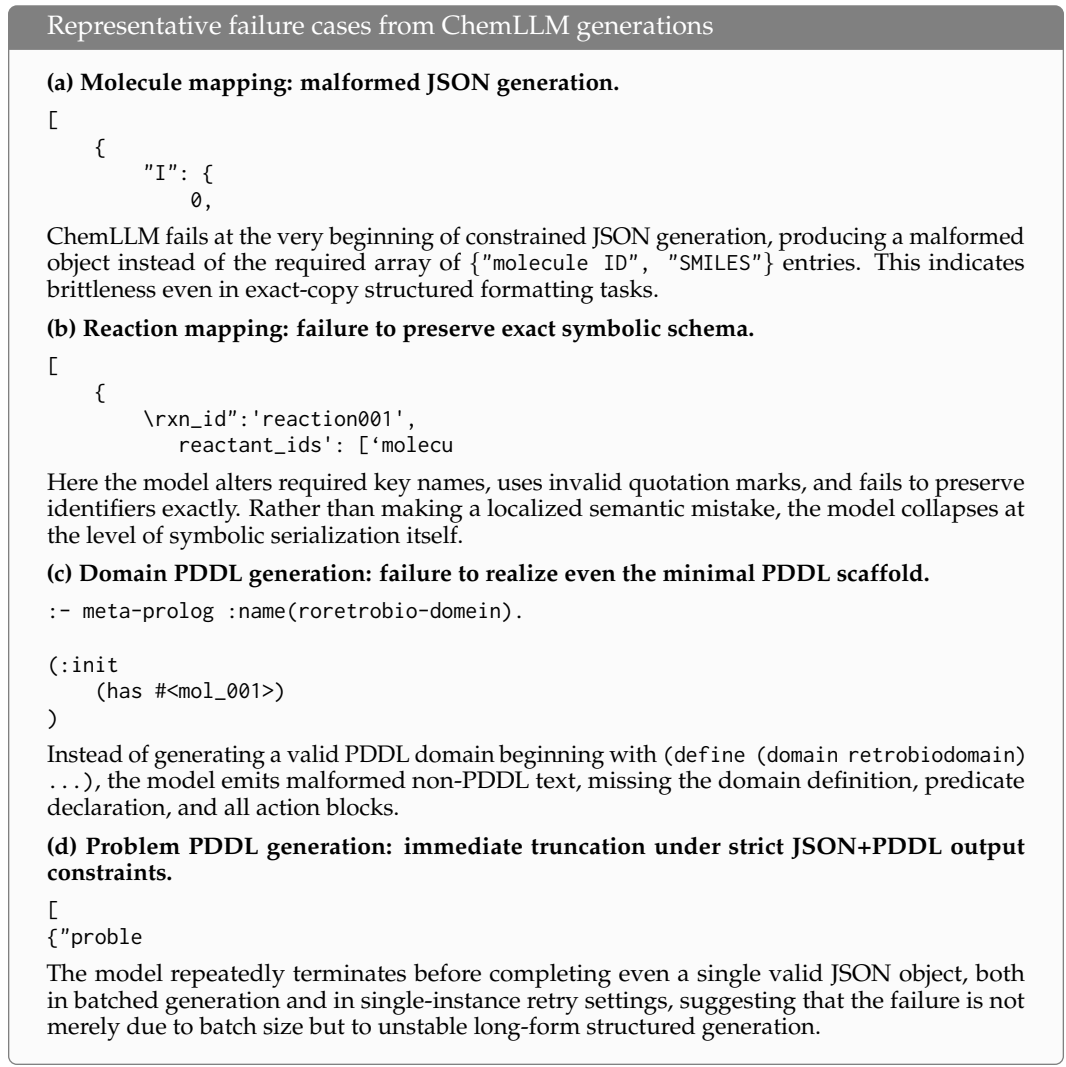

\centering
\begin{tcolorbox}[
    colback=gray!3,
    colframe=black!50,
    width=\textwidth,
    boxrule=0.5pt,
    title={Representative failure cases from ChemLLM generations}
]
\small

\textbf{(a) Molecule mapping: malformed JSON generation.}
\begin{verbatim}
[
    {
        "I": {
            0,
\end{verbatim}
ChemLLM fails at the very beginning of constrained JSON generation, producing a malformed object instead of the required array of \texttt{\{"molecule ID", "SMILES"\}} entries. This indicates brittleness even in exact-copy structured formatting tasks.

\vspace{0.5em}
\textbf{(b) Reaction mapping: failure to preserve exact symbolic schema.}
\begin{verbatim}
[
    {
        “rxn_id”:'reaction001',
           reactant_ids': [‘molecu
\end{verbatim}
Here the model alters required key names, uses invalid quotation marks, and fails to preserve identifiers exactly. Rather than making a localized semantic mistake, the model collapses at the level of symbolic serialization itself.

\vspace{0.5em}
\textbf{(c) Domain PDDL generation: failure to realize even the minimal PDDL scaffold.}
\begin{verbatim}
:- meta-prolog :name(roretrobio-domein).

(:init
    (has #<mol_001>)
)
\end{verbatim}
Instead of generating a valid PDDL domain beginning with \texttt{(define (domain retrobiodomain) ...)}, the model emits malformed non-PDDL text, missing the domain definition, predicate declaration, and all action blocks.

\vspace{0.5em}
\textbf{(d) Problem PDDL generation: immediate truncation under strict JSON+PDDL output constraints.}
\begin{verbatim}
[
{"proble
\end{verbatim}
The model repeatedly terminates before completing even a single valid JSON object, both in batched generation and in single-instance retry settings, suggesting that the failure is not merely due to batch size but to unstable long-form structured generation.

\end{tcolorbox}
\caption{Representative failure modes of ChemLLM across intermediate mapping and symbolic planning tasks. Unlike stronger general-purpose models, ChemLLM frequently fails before downstream planning can be meaningfully evaluated, due to malformed JSON, exact-copy failures, non-PDDL outputs, and truncated generations. These results suggest that chemistry specialization alone is insufficient for executable symbolic planning without robust representation alignment and structured serialization ability.}
\label{fig:chemllm_failure_cases}
\end{figure*}

Overall, these examples suggest that ChemLLM fails upstream of planning. That is, the dominant issue is not simply incorrect chemistry, but the inability to maintain exact symbolic identities and rigid formal structures across constrained generations. This observation reinforces our broader claim that successful retrosynthesis planning depends not only on domain familiarity, but also on robust representation alignment across heterogeneous symbolic formats.



\section{Detailed Experimental Results}
\label{sec:appendix_detailed_results}

This appendix provides the comprehensive dataset and performance metrics for each sub-task in the RetroPlan-Bench across varying dataset scales (100–1000 reactions).

\subsection{Molecule and Reaction Mapping Integrity}
Tables~\ref{tab:molecule_smiles_mapping_integrity} and~\ref{tab:reaction_id_mapping_integrity} summarize the initial stages of representation alignment. 
\begin{itemize}
    \item \textbf{Molecule Mapping (Table~\ref{tab:molecule_smiles_mapping_integrity}):} Stronger models like GPT-5.2 and DeepSeek V3.2~\citep{deepseekai2025deepseekv32pushingfrontieropen} maintain near-perfect identifier match rates, but for other models, SMILES string exact-match rates exhibit a noticeable decay as the number of molecules increases. ChemLLM~\citep{zhang2024chemllmchemicallargelanguage} consistently failed to provide valid outputs in this stage.
    \item \textbf{Reaction ID Mapping (Table~\ref{tab:reaction_id_mapping_integrity}):} Most models demonstrate high robustness in preserving the relational structure of reaction identifiers, indicating that the primary bottleneck is not relational grouping but the underlying chemical string preservation.
\end{itemize}

\subsection{PDDL Generation Validity and Completeness}
The success of symbolic planning depends on the syntactic and semantic correctness of the generated PDDL files.
\begin{itemize}
    \item \textbf{Syntactic Validity (Tables~\ref{tab:pddl_syntax_validity} and~\ref{tab:pddl_domain_validity_completion}):} While GPT-5.2, DeepSeek V3.2~\citep{deepseekai2025deepseekv32pushingfrontieropen}, and the Gemini family maintain high validity across most scales, Qwen models~\citep{qwen2025qwen25technicalreport, yang2025qwen3technicalreport} suffer from a significant breakdown in structural completeness and syntactic correctness as the reaction count exceeds 500.
    \item \textbf{Action Grounding (Table~\ref{tab:pddl_domain_grounding}):} This metric highlights the semantic gap. Even if a domain is syntactically valid, the accuracy of reactant-product grounding decreases with scale, particularly for Gemini 3.1 at the 600-reaction mark.
    \item \textbf{Problem Coverage (Table~\ref{tab:pddl_problem_coverage_100_400}):} GPT-5.2 and DeepSeek V3.2~\citep{deepseekai2025deepseekv32pushingfrontieropen} achieve perfect coverage for problem instances, ensuring that initial and goal states are correctly specified for the planner.
\end{itemize}

\subsection{Retrosynthesis Planning Performance}
The final planning execution results confirm the effectiveness of staged decomposition.
\begin{itemize}
    \item \textbf{Planning Success (Table~\ref{tab:retrosynthesis_results}):} In the staged setting, GPT-5.2, DeepSeek V3.2~\citep{deepseekai2025deepseekv32pushingfrontieropen}, and Gemini 3.1 achieve a 100\% solve rate with high path accuracy. However, Gemini 2.5 Flash fails to solve the problems despite high sub-task accuracy, indicating that even minor cumulative errors prevent executability.
    \item \textbf{Direct Generation Failure (Table~\ref{tab:retrosynthesis_direct_smiles}):} Under the end-to-end (direct SMILES-to-PDDL) setting, all models exhibit a 0\% success rate. This stark contrast emphasizes that the primary challenge is representation alignment rather than raw planning capability.
\end{itemize}


\begin{table*}[t]
\centering
\small
\setlength{\tabcolsep}{6pt}
% \renewcommand{\arraystretch}{1.15}
\resizebox{\textwidth}{!}{
\begin{tabular}{lcccccccccc}
\toprule
\textbf{Model} & \textbf{100} & \textbf{200} & \textbf{300} & \textbf{400} & \textbf{500} & \textbf{600} & \textbf{700} & \textbf{800} & \textbf{900} & \textbf{1000} \\
\midrule
Qwen3-30B-A3B-Thinking-2507
& 0.9913 (0.1652)
& 0.9954 (0.0868)
& --
& --
& 0.9896 (0.0329)
& --
& --
& --
& --
& -- \\
Qwen2.5-14B-Instruct-1M 
& 0.9913 (0.9304)
& --
& --
& --
& --
& --
& --
& --
& --
& -- \\
DeepSeek V3.2 
& 1.0 (1.0)
& 0.9954 (0.6256)
& 1.0 (0.9907)
& 1.0 (0.9928)
& 1.0 (0.9931)
& 0.9987 (0.1797)
& 1.0 (0.9800)
& 1.0 (0.9839)
& 0.9982 (0.1238)
& 1.0 (1.0) \\
GPT-5.2 
& 1.0 (1.0)
& 1.0 (0.9909)
& 1.0 (0.9938)
& 1.0 (0.9904)
& 1.0 (0.9913)
& 1.0 (0.9935)
& 1.0 (0.9944)
& 1.0 (0.9930)
& 1.0 (0.9928)
& 1.0 (0.9921) \\
Gemini 2.5 Flash 
& 1.0 (1.0)
& 1.0 (0.9909) 
& 1.0 (0.9907)
& 1.0 (0.9876)
& 1.0 (0.9844)
& --
& 1.0 (0.9789)
& --
& --
& -- \\
Gemini 3.1 
& 1.0 (1.0)
& 1.0 (0.9909) 
& 1.0 (0.9907)
& 1.0 (0.9928)
& --
& 1.0 (0.9961)
& 1.0 (0.9933)
& --
& --
& -- \\
ChemLLM 
& --
& --
& --
& --
& --
& --
& --
& --
& --
& -- \\
\bottomrule
\end{tabular}
}
\caption{
Task 1 (Molecule Mapping) performance across varying dataset sizes. 
Each cell reports identifier exact-match rate and SMILES exact-match rate (in parentheses). 
``--'' indicates invalid outputs.
}
\label{tab:molecule_smiles_mapping_integrity}
\end{table*}



\begin{table*}[t]
\centering
\small
\setlength{\tabcolsep}{6pt}
\renewcommand{\arraystretch}{1.15}
\resizebox{\textwidth}{!}{
\begin{tabular}{lcccccccccc}
\toprule
\textbf{Model} & \textbf{100} & \textbf{200} & \textbf{300} & \textbf{400} & \textbf{500} & \textbf{600} & \textbf{700} & \textbf{800} & \textbf{900} & \textbf{1000} \\
\midrule
Qwen3-30B-A3B-Thinking-2507
& 1.0000
& 1.0000
& 1.0000
& 1.0000
& 1.0000
& 1.0000
& 1.0000
& --
& --
& -- \\
Qwen2.5-14B-Instruct-1M  
& 1.0000
& 1.0000
& --
& --
& --
& --
& --
& --
& --
& -- \\
DeepSeek V3.2 
& 1.0000
& 1.0000
& 1.0000
& 1.0000
& 1.0000
& 1.0000
& 1.0000
& 1.0000
& 1.0000
& 1.0000\\
GPT-5.2 
& 1.0000
& 1.0000
& 1.0000
& 1.0000
& 1.0000
& 1.0000
& 1.0000
& 1.0000
& 1.0000
& 1.0000 \\
Gemini 2.5 Flash 
& 1.0000
& 1.0000
& 1.0000
& 0.9975
& 1.0000
& 1.0000
& 1.0000
& 0.9988
& 1.0000
& 1.0000 \\
Gemini 3.1 
& 1.0000
& 1.0000
& 1.0000
& 1.0000
& 1.0000
& 1.0000
& 1.0000
& 1.0000
& 1.0000
& 1.0000 \\
ChemLLM 
& --
& --
& --
& --
& --
& --
& --
& --
& --
& -- \\
\bottomrule
\end{tabular}
}
\caption{
Task 2 (Reaction ID Mapping) integrity across varying dataset sizes.
Each cell reports the exact mapping rate (EMR), defined as the proportion of reactions for which 
\texttt{RXN\_ID}, \texttt{Reactants\_IDs}, and \texttt{Products\_IDs} exactly match the input.
``--'' indicates missing or invalid outputs (e.g., empty or unparsable model responses).
}
\label{tab:reaction_id_mapping_integrity}
\end{table*}





\begin{table*}[h!]
\centering
\small
% \setlength{\tabcolsep}{8pt}
\renewcommand{\arraystretch}{1.15}
\begin{tabular}{lcccc}
\toprule
\textbf{Model} & \textbf{100} & \textbf{200} & \textbf{300} & \textbf{400} \\
\midrule
GPT-5.2
& 1.0
& 1.0
& 1.0
& 1.0 \\

DeepSeek V3.2
& 1.0
& 1.0
& 1.0
& 1.0 \\

Qwen3-30B-A3B-Thinking-2507
& 0.0
& 0.1176
& 0.2045
&  0.1531\\

Qwen2.5-14B-Instruct-1M  & 1.0 &  1.0 & 1.0  &1.0\\

Gemini 2.5 Flash
& 1.0
& 1.0
& 1.0
&  1.0\\


Gemini 3.1
& 1.0
& 1.0
& 1.0
&  1.0\\

ChemLLM
& --
& --
& --
& -- \\

\bottomrule
\end{tabular}

\caption{
Syntactic validity rate of generated PDDL problems across different dataset sizes.
A generation is considered valid if it satisfies basic structural constraints, including the correct problem definition, domain declaration, start molecule fact, goal predicates, and balanced parentheses.
}
\label{tab:pddl_syntax_validity}
\end{table*}



\begin{table*}[h!]
\centering
\small
% \renewcommand{\arraystretch}{1.15}
\resizebox{\textwidth}{!}{
\begin{tabular}{llcccccccccc}
\toprule
\textbf{Model} & \textbf{Metric} & \textbf{100} & \textbf{200} & \textbf{300} & \textbf{400} & \textbf{500} & \textbf{600} & \textbf{700} & \textbf{800} & \textbf{900} & \textbf{1000} \\
\midrule

\multirow{2}{*}{GPT-5.2}
& Coverage & 1.0 & 1.0 & 1.0 & 1.0 & 1.0 & 1.0 & 1.0 & 1.0 & 1.0 & 1.0 \\
& Validity & 1.0 & 1.0 & 1.0 & 1.0 & 1.0 & 1.0 & 1.0 & 1.0 & 1.0 & 1.0 \\


\multirow{2}{*}{DeepSeek V3.2}
& Coverage & 1.0 & 1.0 & 1.0 & 1.0 & 1.0 & 1.0 & 1.0 & 1.0 & 1.0 & 1.0 \\
& Validity & 1.0 & 1.0 & 1.0 & 1.0 & 1.0 & 1.0 & 1.0 & 1.0 & 1.0 & 1.0 \\

\multirow{2}{*}{Qwen3-30B-A3B-Thinking-2507}
& Coverage & 1.0 & 1.0 & 0.9800 & 0.7960 & 0.5940 & 0 & 0 & 0 & 0.3380 &0 \\
& Validity & 1.0 & 1.0 & 0 & 0 & 0 & 0 & 0 & 0 & 0 & 0 \\


\multirow{2}{*}{Qwen2.5-14B-Instruct-1M}
& Coverage & 1.0 & 0.8200 & 0.5467 & 0.4131 & 0.3280 & 0.2733 & 0.2343 & 0.2056 & 0.1822 & 0.1600 \\
& Validity & 1.0 & 0 & 0  & 0 & 0 & 0 & 0 & 0 & 0 & 0 \\


\multirow{2}{*}{Gemini 2.5 Flash}
& Coverage & 1.0 & 1.0 & 1.0 & 1.0 & 1.0 & 0.0783 & 1.0 & 1.0 & -- &1.0 \\
& Validity & 1.0 & 1.0 & 1.0 & 1.0 & 1.0 & 0.0 & 1.0 & 1.0 & -- &1.0 \\


\multirow{2}{*}{Gemini 3.1}
& Coverage & 1.0 & 1.0 & 1.0 & 1.0 & 1.0 & 0.0783 & 1.0 & 1.0 & 1.0 &1.0 \\
& Validity & 1.0 & 1.0 & 1.0 & 1.0 & 1.0 & 0.0 & 1.0 & 1.0 & 1.0 &1.0 \\


\multirow{2}{*}{ChemLLM}
& Coverage & -- & -- & -- & -- & -- & -- & -- & -- & -- & -- \\
& Validity & -- & -- & -- & -- & -- & -- & -- & -- & -- & -- \\


\bottomrule
\end{tabular}
}
\caption{
Syntactic validity and action completion rate of generated PDDL domains across different dataset sizes (100–1000 reactions).
Validity indicates whether the generated domain satisfies basic syntactic constraints such as correct domain structure, valid action blocks, and balanced parentheses.
Completion measures the fraction of expected actions successfully generated (e.g., number of generated actions divided by the expected number of reactions).
}
\label{tab:pddl_domain_validity_completion}
\end{table*}



\begin{table*}[h!]
\centering
\small
% \renewcommand{\arraystretch}{1.15}
\resizebox{\textwidth}{!}{
\begin{tabular}{llcccccccccc}
\toprule
\textbf{Model} & \textbf{Metric} & \textbf{100} & \textbf{200} & \textbf{300} & \textbf{400} & \textbf{500} & \textbf{600} & \textbf{700} & \textbf{800} & \textbf{900} & \textbf{1000} \\
\midrule

\multirow{1}{*}{GPT-5.2}
& Grounding &0.9900  &0.9950  &0.9933  &0.9950  &0.9420  &0.8883  &0.8857  &0.8612  &0.8522  &0.8350  \\


\multirow{1}{*}{DeepSeek V3.2}
& Grounding &0.9900  &0.9950  &0.9933  &0.9950  &0.9420  &0.8883  &0.8857  &0.8612  &0.8489  &0.8350  \\


\multirow{1}{*}{Qwen3-30B-A3B-Thinking-2507}
& Grounding &0.9900  &0.9950  &0.9733  &0.7884  &0.5880  &0  &0  &0.3356  &0  &0  \\


\multirow{1}{*}{Qwen2.5-14B-Instruct-1M}
& Grounding &0.9900  &0.7900  &0.5367  &0.3501  &0.3240  &0.2683  &0.2300  &0.2012  &0.1622  &0.1580  \\


\multirow{1}{*}{Gemini 2.5 Flash}
& Grounding &0.9900  &0.9950  &0.9933  &0.9950  &0.9420  &0.8883  &0.8857  &0.8612  &0  &0.8350  \\


\multirow{1}{*}{Gemini 3.1}
& Grounding &0.9900  &0.9950  &0.9933  &0.9950  &0.9420  &0.0767  &0.8857  &0.8612  &0.8522  &0.8350  \\


\multirow{1}{*}{ChemLLM}
& Grounding & 0  &0  &0  &0  &0  &0  &0  &0  &0  &0  \\


\bottomrule
\end{tabular}
}

\caption{
Action grounding performance of generated PDDL domains across different dataset sizes (100–1000 reactions).
Grounding measures whether each generated action correctly includes the corresponding reactant and product identifiers for the given reaction.
This metric evaluates semantic correctness beyond syntactic validity and action coverage.
}
\label{tab:pddl_domain_grounding}
\end{table*}


\begin{table*}[h!]
\centering
\small
\resizebox{\textwidth}{!}{
\begin{tabular}{llcccc}
\toprule
\textbf{Model} & \textbf{Metric} & \textbf{100} & \textbf{200} & \textbf{300} & \textbf{400} \\
\midrule

\multirow{1}{*}{GPT-5.2}
& Problem Coverage &1.0  &1.0  &1.0  &1.0  \\


\multirow{1}{*}{DeepSeek V3.2}
& Problem Coverage &1.0  &1.0  &1.0  &1.0  \\


\multirow{1}{*}{Qwen3-30B-A3B-Thinking-2507}
& Problem Coverage &1.0  &0.8897  &0.8636  &0.8980  \\


\multirow{1}{*}{Qwen2.5-14B-Instruct-1M}
& Problem Coverage &0.1304  &0.3309  &0.5909  &0.5442  \\


\multirow{1}{*}{Gemini 2.5 Flash}
& Problem Coverage &1.0  &1.0  &1.0  &1.0  \\


\multirow{1}{*}{Gemini 3.1}
& Problem Coverage &1.0  &1.0  &1.0  &1.0  \\


\multirow{1}{*}{ChemLLM}
& Problem Coverage &0  &0  &0  &0  \\


\bottomrule
\end{tabular}
}

\caption{
Problem coverage of generated PDDL problem files across dataset sizes (100–400 reactions).
Problem coverage measures the fraction of ground-truth problem instances that are correctly generated, 
where each problem file is aligned with the input molecule (Mol\_Name), and includes both the corresponding 
final product (Final\_Product\_ID) and all required goal candidate molecules.
}
\label{tab:pddl_problem_coverage_100_400}
\end{table*}


\begin{table*}[h!]
\centering
\small
\renewcommand{\arraystretch}{1.15}
\begin{tabular}{lcc}
\toprule
\textbf{Model} & \textbf{Solve Rate} & \textbf{Path Accuracy} \\
\midrule
GPT-5.2 &1.0  &0.9864  \\

DeepSeek V3.2 &1.0  &0.9864  \\

Qwen3-30B-A3B-Thinking-2507 &0.0  &0.0  \\

Qwen2.5-14B-Instruct-1M  &0.0 &0.0 \\

Gemini 2.5 Flash &0  &0  \\

Gemini 3.1 &1.0  &0.9864  \\

ChemLLM & 0  & 0  \\

\bottomrule
\end{tabular}

\caption{
Retrosynthesis planning performance on the 400-problem test set.
Solve Rate indicates the fraction of problems for which the planner successfully generated a valid plan.
Path Accuracy measures whether the predicted reaction sequence matches the ground-truth pathway, allowing both forward and reverse order matches to account for the retrosynthesis direction.
}
\label{tab:retrosynthesis_results}
\end{table*}




\begin{table*}[h!]
\centering
\small
\renewcommand{\arraystretch}{1.15}
\begin{tabular}{lcc}
\toprule
\textbf{Model} & \textbf{Solve Rate} & \textbf{Path Accuracy} \\
\midrule
GPT-5.2 & 0  & 0  \\

DeepSeek V3.2 & 0 & 0 \\

Qwen3-30B-A3B-Thinking-2507 & 0  & 0  \\

Qwen2.5-14B-Instruct-1M & 0  & 0  \\

Gemini 2.5 Flash & 0  & 0  \\

Gemini 3.1 & 0  & 0  \\

ChemLLM & 0  & 0  \\

\bottomrule
\end{tabular}

\caption{
Retrosynthesis planning performance using direct SMILES-based PDDL generation.
In this setting, models are provided with the reactant and product SMILES strings and must directly generate both the \texttt{domain.pddl} and \texttt{problem.pddl} required for planning. 
This experiment evaluates a more direct generation pipeline compared to the reaction-template-based setting, serving as a comparison of end-to-end reasoning capability.
}
\label{tab:retrosynthesis_direct_smiles}
\end{table*}



\appendix
\section{Prompt Examples}


\begin{figure*}[t]
\centering
\begin{tcolorbox}[
    colback=gray!3,
    colframe=black!50,
    width=\textwidth,
    boxrule=0.5pt,
]
\begin{minipage}{\textwidth}
\small
\begin{verbatim}
You are performing a data formatting task for a molecular dataset preprocessing pipeline.

Task:
Given an ordered list of SMILES strings, 
assign molecule identifiers sequentially starting from molecule0001.
Return a JSON array of length {N}. 
Each element must be an object with exactly two keys: 
"MOL_ID" and "SMILES".

Hard constraints:
1) Preserve each SMILES string EXACTLY as provided (character-for-character).
2) Output ONLY a valid JSON array. No extra text, no markdown, no explanations.
3) The output must contain exactly {N} items, in the same order as the input list.

Input SMILES list:
{SMILES_LIST}
\end{verbatim}
\end{minipage}
\end{tcolorbox}
\caption{
Dataset generation prompt for molecular dataset preprocessing.
The prompt assigns sequential molecule identifiers to an ordered list of SMILES strings while
preserving the original string representation exactly and enforcing strict JSON-only output format.
}
\label{fig:smiles_id_mapping_prompt}
\end{figure*}

\begin{figure*}[t]
\centering
\begin{tcolorbox}[
    colback=gray!3,
    colframe=black!50,
    width=\textwidth,
    boxrule=0.5pt,
]
\begin{minipage}{\textwidth}
\small
\begin{verbatim}
You are given {N} chemical reactions.

Each reaction entry contains:
- RXN_ID (already assigned and must be preserved)
- Reactants_IDs
- Products_IDs

Your task is to return the same reactions in a structured JSON format.

Return ONLY a valid JSON array of length {N}.
Each element in the array must be an object with exactly the following keys:
- "RXN_ID"
- "Reactants_IDs"
- "Products_IDs"

The order of the output array must exactly match the order of the input reactions.

Do not change any IDs or values.
Do not add, remove, or reorder any entries.
Do not include any explanations, markdown, or extra text outside the JSON array.

REACTIONS:
{REACTIONS_LIST}
\end{verbatim}
\end{minipage}
\end{tcolorbox}
\caption{
Dataset formatting prompt for reaction-structured preprocessing.
The prompt enforces exact preservation of reaction identifiers and molecule ID sets while converting
an ordered list of reactions into a strict JSON array with fixed keys and original ordering.
}
\label{fig:reaction_json_format_prompt}
\end{figure*}

\begin{figure*}[t]
\centering
\begin{tcolorbox}[
    colback=gray!3,
    colframe=black!50,
    width=\textwidth,
    boxrule=0.5pt,
]
\begin{minipage}{\textwidth}
\small
\begin{verbatim}
You are a PDDL generator for retrosynthesis planning.

Your task:
- Generate multiple valid PDDL problem files for the domain `retrobiodomain`.
- Return ONLY a valid JSON array with exactly {N} elements. No extra text.

For each element i in the JSON array, output an object with exactly these keys:
- "problem_name"
- "start_mol"
- "problem_pddl"

PDDL constraints (must follow exactly):
- The PDDL must use exactly this structure (including the comment):

(define (problem PROBLEM_NAME)
  (:domain retrobiodomain) ; should match the name

  (:init
    (has START_MOLECULE)
  )

  (:goal
    (or
      GOAL_FACTS
    )
  )
)

Where:
- PROBLEM_NAME must be replaced by the provided problem name.
- START_MOLECULE must be replaced by the provided start molecule ID.
- GOAL_FACTS must be replaced by the list of goal molecules, 
each in the form (has MOLxxxxx) on its own line, properly indented.

Hard rules:
- Do NOT invent any new molecule IDs.
- Do NOT rename the domain name `retrobiodomain`.
- Do NOT add any predicates, functions, types, costs, metrics, or constraints.
- Keep output strictly as PDDL problem definitions, with no explanations.

Input format:
- You will be given:
  1) A list of problems with (problem_name, start_mol)
  2) A single shared goal list (already provided as molecule IDs)

Problems:
{PROBLEMS_LIST}

Shared goals:
{GOALS_LIST}
\end{verbatim}
\end{minipage}
\end{tcolorbox}
\caption{
Prompt for generating multiple PDDL problem instances for retrosynthesis planning.
Each problem specifies an initial molecule and shared goal conditions, enforcing a strict
structure compatible with the \texttt{retrobiodomain} domain and standard PDDL planners.
}
\label{fig:pddl_problem_generation_prompt}
\end{figure*}

\begin{figure*}[t]
\centering
\begin{tcolorbox}[
    colback=gray!3,
    colframe=black!50,
    width=\textwidth,
    boxrule=0.5pt,
]
\begin{minipage}{\textwidth}
\small
\begin{verbatim}
You are writing a PDDL domain file for classical planning 
to be run on a standard PDDL planner using the LAMA-first planner.

You will be given a list of chemical reactions.
Each reaction provides:

- RXN_ID
- Reactants_IDs (one or more molecule IDs)
- Products_IDs (one or more molecule IDs)

Your job:

- Output ONLY a valid PDDL domain definition (domain.pddl). 
No explanations. No markdown.
- The domain name must be exactly: retrobiodomain
- Use a single predicate: (has ?m)
- For each reaction, create one action with the same name as RXN_ID.
- Interpret each action in retrosynthesis direction.
- Therefore:

  - Preconditions: require all product molecule IDs as (has P1) (has P2) ...
  - Effects: add all reactant molecule IDs as (has R1) (has R2) ...
  - Also delete all product molecule IDs using (not (has Pi))
- Do not invent molecule IDs or reactions.
- Do not add any additional predicates, functions, costs, metrics, or constraints.
- Preserve the reaction order exactly as given.

Important formatting rules:

- Use only PDDL constructs supported by common planners.
- Keep it simple (STRIPS-style actions with add/delete effects).
- If a reaction has multiple product molecule IDs, 
all of them must appear in the precondition.
- If a reaction has multiple reactant molecule IDs, 
all of them must be added in the effect.
- Do not include types or parameters beyond ?m in the predicate declaration.
- The output must be directly usable as a domain.pddl file.

REACTIONS (preserve order):
{REACTIONS_LIST}
\end{verbatim}
\end{minipage}
\end{tcolorbox}
\caption{
Prompt for generating a PDDL domain file for retrosynthesis planning.
The prompt converts an ordered list of chemical reactions into planner-compatible STRIPS-style actions
while preserving reaction identifiers, molecule identifiers, and the original reaction order.
}
\label{fig:pddl_domain_generation_prompt}
\end{figure*}

\end{document}